\pdfoutput=1
\documentclass[11pt,a4paper]{article}
\usepackage[final]{papertwocol}
\usepackage{times}
\usepackage{latexsym}
\usepackage[T1]{fontenc}
\usepackage[utf8]{inputenc}
\usepackage{microtype}
\usepackage{inconsolata}
\usepackage{graphicx}
\usepackage{amsmath}
\usepackage{booktabs}
\usepackage{array}
\usepackage{multirow}
\usepackage{placeins}
\usepackage{helvet,tabularx,colortbl}
\definecolor{headgray}{gray}{0.93}
\newcommand{\diffval}[1]{{\bfseries\boldmath #1}}
\newcommand{\grouprule}{\specialrule{.55pt}{0pt}{0pt}}
\title{CogBattery: Human-Referenced Evaluation of Cognitive Task Behavior and Structure in Language Models}
\author{Dengzhe Hou\textsuperscript{1,*}\qquad Lingyu Jiang\textsuperscript{1}\qquad Fangzhou Lin\textsuperscript{1,2,3}\qquad Kazunori D Yamada\textsuperscript{1}\\[3pt]
{\normalfont \textsuperscript{1}Tohoku University\qquad \textsuperscript{2}Texas A\&M University\qquad \textsuperscript{3}Worcester Polytechnic Institute}\\[3pt]
{\normalfont\small \textsuperscript{*}Corresponding author.\quad Project: \href{https://github.com/dengzhe-hou/CogBattery}{\texttt{github.com/dengzhe-hou/CogBattery}}}}
\date{}
\begin{document}
\raggedbottom
\maketitle

\begin{abstract}
Human cognitive tasks let us compare language models with humans under controlled conditions. Yet a model can reproduce the direction of a human effect while making different errors. CogBattery evaluates 54 open-weight language models on text adaptations of 13 cognitive paradigms, comparing each with human data from its source study. Like humans, models respond more to n-back lures than to controls, yet they make far more false alarms on ordinary non-targets. Most models' reported confidence separates correct from incorrect answers less well than humans' does, and most always choose the same wager, including models that classify accurately. These patterns span many model families. Across tasks, model scores share two dimensions. The first is closely tied to scale, yet larger models are not consistently closer to humans. Predefined cognitive groupings show no reliable predictive advantage over random groupings of the same size. These results show that matching human effect directions does not establish human-like behavior. CogBattery provides condition-level human references for testing errors and confidence reports in new models.
\end{abstract}

\begin{figure*}[t]
\centering
\setlength{\abovecaptionskip}{3pt}
\setlength{\belowcaptionskip}{-10pt}
\includegraphics[width=\textwidth]{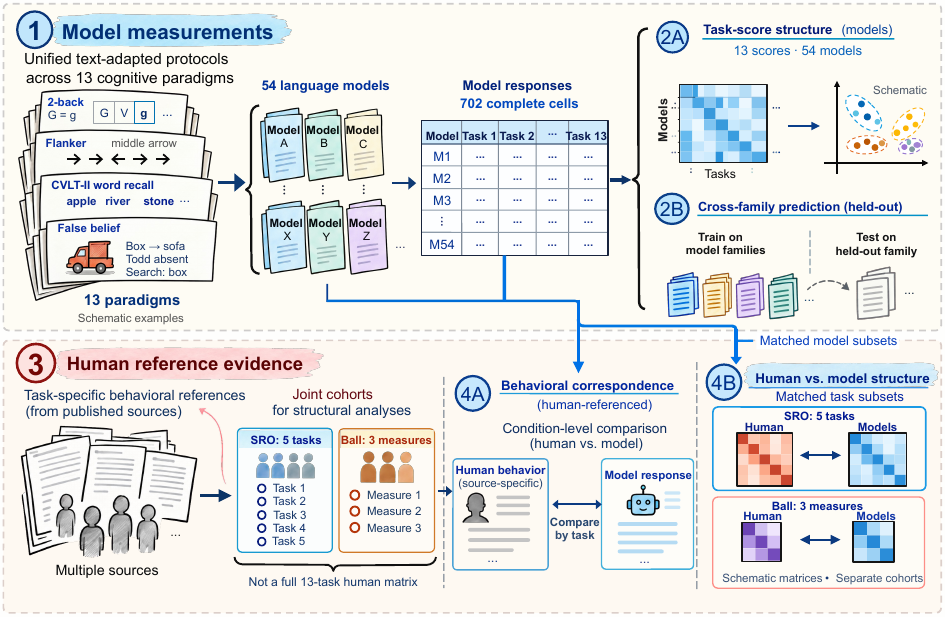}
\caption{Overview of CogBattery. Panel 2B tests whether task groups add value in family-held-out prediction.}
\label{fig:h54-overview}
\end{figure*}

\section{Introduction}
Human cognitive tasks compare conditions that differ in a controlled way, and published human data show how people respond in each. They therefore allow direct comparisons between language models and humans and reveal errors hidden by aggregate accuracy \citep{codaforno2024cogbench,binz2023triangulating,momente2025triangulating}.

A model that shows a human condition effect, such as a decision bias \citep{binz2023triangulating,itzhak2024instructed}, can look human-like. But matching the direction of an effect does not mean making the same errors or reports. A model's confidence can be higher on correct than on incorrect answers, as people's is, yet separate the two less well. Studies of decision making, n-back, and executive function show that shared effects can mask different errors, sensitivities, and confidence reports \citep{codaforno2024cogbench,hu-lewis-2025-language,delangis2026strong}, but each covers few paradigms or few models (Table~\ref{tab:h54_related}), so how general these differences are remains unclear.

CogBattery evaluates 54 open-weight models from 23 families on text adaptations of 13 paradigms (Figure~\ref{fig:h54-overview}). No human cohort covers all 13 paradigms, so each paradigm is compared with source-specific published human data, mostly participant-level records, by condition and response type (Table~\ref{tab:h54_task_guide}). We also check whether differences hold among models that reach a minimal level on each task and whether they vary with model size. The paradigms fall into five groups by origin, such as working memory and metacognition, and we test whether these groups predict held-out families' task scores better than random groups of equal size. Task correlations need the same people on several tasks, and robust condition effects do not guarantee reliable individual differences \citep{hedge2018reliability}, so structure is compared only in four such cohorts \citep{eisenberg2019ontology,ball2022monitoring,poldrack2016phenome,bradford2020belief}.

Our findings are as follows. First, shared directions hide different errors. The n-back lure effect sits on many false alarms to ordinary non-targets, an excess of .257 over humans, and Stroop interference is much larger than humans' (Figure~\ref{fig:h54-behavior}, Table~\ref{tab:h54_behavior_core}). Second, verbal confidence separates correct from incorrect answers less well than humans' does, although token probabilities do about as well, and most models always choose the same wager. Third, several gaps remain among gate passers, and the n-back excess remains in models of 27B or more (Table~\ref{tab:h54_behavior_gates}). Fourth, model scores share a scale-related dimension, yet larger models are not consistently closer to humans (Figure~\ref{fig:h54-closeness}), and predefined groups add no reliable predictive value over random groups.

Comparisons with humans thus need errors and confidence reports, not only effect directions. CogBattery enables such comparisons for new models, with the same text materials, condition-level human references, and scorers used here.

\begingroup
\raggedbottom
\section{Related Work}
\paragraph{Cognitive tasks and human-referenced behavior.}
Several evaluations port cognitive tasks to language models. CogBench separates performance from behavior \citep{codaforno2024cogbench}, CogEval tests cognitive claims with controls \citep{momennejad2023cogeval}, and triangulation links cognitive tests, games, and benchmarks \citep{momente2025triangulating}. Others predict human choices directly \citep{binz2025centaur}. Single-paradigm studies show that instruction tuning can strengthen human-like decision biases \citep{itzhak2024instructed}. Working-memory and executive-function studies probe capacity and control \citep{gong2024working,delangis2026strong}, n-back failures can reflect task misunderstanding rather than capacity \citep{hu-lewis-2025-language}, and cumulative state tracking ties working-memory difficulty to the number of sequential updates \citep{hou2026wmfam}. Direct human comparisons find differences within shared effect directions. Visual Stroop shows different error curves \citep{patel2026deficient}, confidence differences depend on the task \citep{cash2026quantifying}, and GPT-4 has lower AUROC (area under the receiver operating characteristic curve) than humans on four-choice but not two-choice Japanese general knowledge \citep{yoshizawa2026metacognition}. Some models use confidence to guide abstention \citep{kumaran2026causal}.

The evaluations in Table~\ref{tab:h54_related} cover few paradigms or few models, and they use human data only as norms or scaling constants, if at all. CogBattery asks how widely specific errors and confidence reports recur, comparing 54 models from 23 families with human data from each paradigm's source study. Text adaptations and auxiliary demands constrain interpretation \citep{hu2024auxiliary}, and protocol differences preclude matched comparisons to published model scores.

\begin{table*}[t]
\centering
\fontsize{8.5}{9.6}\selectfont
\setlength{\tabcolsep}{1.7pt}
\begin{tabular}{@{}lrr>{\raggedright\arraybackslash}p{.20\textwidth}c>{\raggedright\arraybackslash}p{.29\textwidth}c@{}}
\toprule
Study & \shortstack{Tasks/\\abilities} & Models & Human reference & \shortstack{Errors vs.\ humans\\by condition} & \shortstack[l]{Cross-task\\association/embedding} & \shortstack{Family-held-out\\prediction} \\
\midrule
CogBench & 7 & 40 & Means, scaling only & $\circ$ & $\bullet$ UMAP & $\circ$ \\
CogEval & 1 & 8 & None & $\circ$ & $\circ$ & $\circ$ \\
Momentè & 11 & 9 & None & $\circ$ & $\bullet$ Rank correlations & $\circ$ \\
de Langis & 7 & 10 & Published norms & $\bullet$ & $\bullet$ WM--EF correlation & $\circ$ \\
Hu--Lewis & 1 & 8 & None & $\circ$ & $\circ$ & $\circ$ \\
CogProbe & 16 & 4 & None & $\circ$ & $\circ$ & $\circ$ \\
\rowcolor{headgray}
\textbf{CogBattery} & \textbf{13} & \textbf{54} & Each task's source study, mostly individual records & $\bullet$ & $\bullet$ Correlations, PCA, factor analysis & $\bullet$ \\
\bottomrule
\end{tabular}
\caption{Closest cognitive-task evaluations \citep{codaforno2024cogbench,momennejad2023cogeval,momente2025triangulating,delangis2026strong,hu-lewis-2025-language,cui2026cogprobe}. Filled circles mark reported analyses and open circles unreported ones. Family-held-out prediction tests families absent from training. Human and model task correlations are compared in jointly measured subsets. CogBench's count follows the proceedings.}
\label{tab:h54_related}
\end{table*}

\paragraph{Profiles and task-score structure.}
Profiling systems use cognitive labels \citep{zhou2026adele,haznitrama2026neuro,cui2026cogprobe}. Our five groups denote paradigm origins in working memory, cognitive control, episodic memory, social reasoning, and metacognition \citep{miyake2000unity,wellman2001tom,lichtenstein1977calibration}. Benchmark factor studies motivate testing shared variation, scale, and incremental prediction \citep{burnell2023revealing,ilic2024evidence,kipnis2025metabench}. Condition effects do not guarantee reliable individual differences \citep{hedge2018reliability}, and task convergence and construct validity require separate evidence \citep{trott2026tomtasks,bean2025measuring,jung2026psychometric}. We therefore test whether these paradigm-origin groups add predictive value for task scores in held-out model families.

\endgroup
\section{Battery and Administration}
\label{sec:h54_design}
\subsection{Tasks and human references}
\begin{table*}[!t]
\centering
\fontsize{8.5}{9.4}\selectfont
\setlength{\tabcolsep}{3pt}
\setlength{\extrarowheight}{.5pt}
\setlength{\arrayrulewidth}{.25pt}
\setlength{\abovecaptionskip}{3pt}
\begin{tabular}{>{\raggedright\arraybackslash}p{.155\textwidth}>{\raggedright\arraybackslash}p{.235\textwidth}>{\raggedright\arraybackslash}p{.22\textwidth}>{\raggedright\arraybackslash}p{.33\textwidth}}
\specialrule{.6pt}{0pt}{0pt}
\rowcolor{headgray}
Task & What the model does & Measures (human source) & Changes for models \\
\specialrule{.6pt}{0pt}{0pt}
\multicolumn{4}{l}{\textbf{Working memory}} \\
\hline
\textbf{1} Digit span (DS) & Repeat digits forward or backward. & Sequence accuracy by length and direction (\textit{SRO}) & \textit{Visible.} \textbf{No} timing or adaptive length. \\
\hline
\textbf{2} N-back (NB) & Check whether the current letter matches 2 or 3 steps back. & MATCH rates for targets, lures, other non-targets (\textit{SRO}) & \textit{Visible.} MATCH/NO MATCH replaces keys.\newline\textbf{No} timing, practice or adaptive load. \\
\hline
\textbf{3} Operation span (OS) & Judge equations, then recall letters. & Letter accuracy by position (\textit{Ball}) & \textit{Visible.} Equations from public pools.\newline\textbf{No} processing deadline. \\
\specialrule{.6pt}{0pt}{0pt}
\multicolumn{4}{l}{\textbf{Cognitive control}} \\
\hline
\textbf{4} Stroop (ST) & Name the color label and ignore the word. & Accuracy with and without conflict (\textit{SRO}) & \textit{Untimed.} Color labels replace ink.\newline\textbf{No} reaction time. \\
\hline
\textbf{5} Flanker (FL) & Find the center symbol among matching, conflicting or neutral neighbors. & Accuracy by condition and cue (\textit{SRO}) & \textit{Untimed.} Text arrows and cues, LEFT/RIGHT answers.\newline\textbf{No} cue timing, deadline or reaction time. \\
\hline
\textbf{6} Go/no-go (GN) & PRESS for the GO color, otherwise WITHHOLD. & Hits, false alarms, $d'$ (\textit{SRO}) & \textit{Untimed.} Color words, recent history visible.\newline\textbf{No} response window. \\
\specialrule{.6pt}{0pt}{0pt}
\multicolumn{4}{l}{\textbf{Episodic memory}} \\
\hline
\textbf{7} CVLT-II & Recall list words repeatedly, then recognize them. & Recall, intrusions (unlisted words), recognition (\textit{CNP}) & \textit{Visible.} Written list stays visible.\newline\textbf{No} real delays, only stage labels. \\
\hline
\textbf{8} DRM & Recognize studied words, related lures and unrelated words, with or without warning. & Hits and false alarms by word type (\textit{Ball}) & \textit{Visible.} With and without warning per model, one fixed shuffle, study words visible.\newline\textbf{No} delay. \\
\hline
\textbf{9} Source memory (SM) & Identify who provided information in quoted stories. & Source accuracy by quotation style (\textit{Eerland}) & \textit{Visible.} Three stories per model, two fixed question forms, stories visible. \\
\specialrule{.6pt}{0pt}{0pt}
\multicolumn{4}{l}{\textbf{Social reasoning}} \\
\hline
\textbf{10} False belief (FB) & Answer location questions in false- and true-belief stories. & Accuracy on belief and check questions (\textit{Strachan}) & \textit{Same items.} Fixed scoring rules, both lists per model. \\
\hline
\textbf{11} Recursive mindreading (EP) & Answer mental-state and control questions at levels 1--7. & Accuracy by level and type (\textit{EPITOME}) & \textit{Same items.} A/B answers, all format combinations per model.\newline\textbf{No} confidence ratings. \\
\specialrule{.6pt}{0pt}{0pt}
\multicolumn{4}{l}{\textbf{Metacognition}} \\
\hline
\textbf{12} Calibration (CAL) & Answer French two-choice questions and rate confidence. & Accuracy, confidence bias, type-2 AUROC (\textit{Mazancieux}) & \textit{Same items.} Each item in both answer orders. \\
\hline
\textbf{13} Wagering (WAG) & Classify grammar strings and choose a low or high wager. & Accuracy, high-minus-low gap if defined (\textit{Dienes}) & \textit{Visible.} Copy visible strings, both grammars and presentation orders.\newline\textbf{No} real sweets. \\
\specialrule{.6pt}{0pt}{0pt}
\end{tabular}
\caption{The 13 paradigms and model adaptations. All model responses are text. CVLT is the California Verbal Learning Test, and DRM is the Deese--Roediger--McDermott paradigm. Italic tags mark the main change. \textit{Visible} means materials or history stay in context, \textit{Untimed} means human timing and motor demands are omitted, and \textit{Same items} means the human study used the same written items. Other conditions are kept. Human cohorts and item correspondence vary by source.}
\label{tab:h54_task_guide}
\end{table*}

Table~\ref{tab:h54_task_guide} defines the tasks and their five groups, and Table~\ref{tab:h54_structural_scores} gives the score of each task used in the structure analyses. Each task is compared with human data from its source study, and most of these are participant-level records. For DS, ST, FL, and GN the models replay three public trial sequences of Self-Regulation Ontology (SRO) participants \citep{eisenberg2019ontology}, and NB uses 80 participant-matched blocks. CVLT uses 130 healthy Consortium for Neuropsychiatric Phenomics (CNP) controls on the CVLT-II Standard lists, with all 272 records in a sensitivity analysis \citep{delis2000cvlt,poldrack2016phenome}. Table~\ref{tab:h54_task_guide} names the sources for the other tasks \citep{ball2022monitoring,eerland2018source,strachan2024tom,jones2024epitome,mazancieux2020g}, and each CAL item appears in both answer orders. Two references are weaker. OS reconstructs Ball's rules with different item orders, and WAG has only published grammar strings and human summaries \citep{dienes2005measuring,dienes2010gambling}.

\paragraph{Example item.}
For 2-back, \texttt{1:G 2:V 3:g} requires \texttt{MATCH} because the current letter matches two positions earlier, ignoring case. Earlier letters and the model's earlier answers stay visible.

\subsection{Model panel and execution}
The 54 open-weight models span 23 families, merging versions of the same series (28 raw labels), and 0.36B--141B total parameters, including all mixture-of-experts parameters. Families include Qwen, Llama, Gemma, and Mistral. Sensitivity panels retain a prespecified 19 models in nine families or the 51 direct-answer models in 22 families.

All models share prepared tasks and validity rules. Qwen3 4B and DeepSeek-R1 (7B and 14B) use native thinking. The other 51 answer directly. Temperature is zero, with one response per step. Formats are constrained except free-response FB. Models are served with Ollama 0.18.0. Contexts are 4,096 tokens, or 12,288 with 8,192 output tokens for native thinking, and OS allows 128 output tokens for every model.

\subsection{Response validity and behavioral coverage}
All 702 cells completed, giving 366,606 records including non-outcome steps. Of these, 366,248 are valid and scorable, 120 are technically invalid, 238 are valid but unparsed, and none were left unrun, so 366,486 (99.97\%) are technically valid. Technical validity concerns response format and delivery, not behavioral correctness. The 238 unparsed replies are FB answers without an extracted location, and they count as wrong.

Invalid task answers count as wrong, including misses on signal trials and false alarms on noise trials, without a coverage gate. NB has no invalid answers, so its false-alarm rates do not depend on this rule. Confidence and wager reports still require validity because they lack wrong-answer codes. Type-2 AUROC, the area under the ROC curve for confidence separating correct from incorrect answers, requires both answer outcomes, and high-minus-low accuracy requires both wager levels. Otherwise each remains undefined. Signal-detection $d'$ adds .5 to event counts and 1 to trial counts and clips rates to [.001,.999]. Missing replies, hard errors, and context overflow end a chain without regeneration.

\section{Analysis}
\label{sec:h54_analysis}
\begin{table*}[!t]
\centering
\fontsize{8.5}{9.0}\selectfont
\setlength{\tabcolsep}{3pt}
\setlength{\extrarowheight}{0pt}
\setlength{\arrayrulewidth}{.25pt}
\setlength{\abovecaptionskip}{3pt}
\begin{tabular}{>{\raggedright\arraybackslash}p{.04\textwidth}>{\raggedright\arraybackslash}p{.23\textwidth}>{\raggedright\arraybackslash}p{.24\textwidth}>{\raggedright\arraybackslash}p{.43\textwidth}}
\specialrule{.6pt}{0pt}{0pt}
\rowcolor{headgray}
No. & Task & Score used & Readout and aggregation \\
\specialrule{.6pt}{0pt}{0pt}
\textbf{1} & Digit span (DS) & Digit-position accuracy & Across requested digit positions. \\
\hline
\textbf{2} & N-back (NB) & Block $d'$ & Average blocks within each load, then weight 2-back and 3-back equally. \\
\hline
\textbf{3} & Operation span (OS) & Letter-position accuracy & Across requested letter positions. \\
\hline
\textbf{4} & Stroop (ST) & Condition-balanced accuracy & Equal weight for congruent and incongruent conditions. \\
\hline
\textbf{5} & Flanker (FL) & Condition-balanced accuracy & Equal weight for congruent, incongruent, and neutral conditions. \\
\hline
\textbf{6} & Go/no-go (GN) & $d'$ & Discrimination from hit and false-alarm rates. \\
\hline
\textbf{7} & CVLT-II & Total recall & Sum recall over the five learning rounds. \\
\hline
\textbf{8} & DRM & Class-balanced accuracy & Average class-balanced scores with and without warning. \\
\hline
\textbf{9} & Source memory (SM) & Source accuracy & Across the three quotation conditions. \\
\hline
\textbf{10} & False belief (FB) & Main-item accuracy & The 14 main items in one conversation, with a system message. \\
\hline
\textbf{11} & Recursive mindreading (EP) & Mental-question accuracy & Equal weight for orders 1--7. \\
\hline
\textbf{12} & Calibration (CAL) & Type-2 AUROC & Use confidence to discriminate correct and incorrect answers. \\
\hline
\textbf{13} & Wagering (WAG) & Type-2 AUROC & Use low or high wagers to discriminate correct and incorrect answers. \\
\specialrule{.6pt}{0pt}{0pt}
\end{tabular}
\caption{One score per task forms the 54-model by 13-task matrix used in the main structure analyses. Task numbers match Table~\ref{tab:h54_task_guide}, and higher is better for all scores. AUROC treats higher confidence or wagers as predicting correctness, and constant wagers with both correct and incorrect answers give .5.}
\label{tab:h54_structural_scores}
\end{table*}

\subsection{Behavioral endpoints and family coverage}
This step asks how far models are from humans on each task's own measures. Endpoints follow each source study's conditions and response types (Table~\ref{tab:h54_task_guide}). For calibration we report accuracy, confidence, the source study's bias measure, and type-2 AUROC computed in one fixed direction. Contrasts that cannot be computed stay missing.

For endpoint $e$, the primary estimate is the equal-weight mean of $x_{me}-\bar h_e$ over eligible models. Its descriptive 95\% interval resamples whole families 5,000 times, keeping the members of each sampled family and the human mean $\bar h_e$ fixed. Sensitivities weight families equally, use the 28 raw family labels, restrict to the 19- or 51-model panels, or delete one family at a time. Counts of models with a given sign describe the panel and are not significance tests. For Figure~\ref{fig:h54-closeness} we divide each model's absolute difference $|x_{me}-\bar h_e|$ by the across-person human SD of $e$. SM uses the pooled within-condition SD, EP the SD of per-person slopes, and WAG the published SE times $\sqrt{40}$ for high-minus-low accuracy.

\subsection{Structure within common human samples}
This step asks whether tasks from the same paradigm group correlate more strongly than tasks from different groups, in humans and in models. No human cohort covers all 13 tasks, but two cohorts did several within-group tasks. SRO supplies 345 people with DS, NB, ST, FL, and GN, and Ball supplies 409 with OS, DRM, and SM (236 unwarned, 173 warned). Ball SM matches the model task only at task-family level. CNP controls (CVLT, DS, ST) and Bradford adults (FB, OS, ST) have one task per group, so for them we compare each task pair's correlation.

In SRO and Ball, and in model scores, $\delta$ is the mean Pearson correlation of within-group task pairs minus that of between-group pairs.
Larger $\delta$ means stronger relative grouping, not higher ability. SRO has 4 within-group and 6 between-group pairs, and Ball has 1 and 2. Human-minus-model differences in $\delta$ use independent 5,000-draw bootstraps over people and over model families. Missing human and invalid model answers count as wrong.

\subsection{Model dimensions and prediction}
\label{sec:h54_fullbattery_methods}
This step asks how many dimensions model scores share and whether paradigm groups predict scores in unseen model families. One structural score per task (Table~\ref{tab:h54_structural_scores}) gives a 13-column matrix for all 54 models in 23 families. Key endpoints are often contrasts with mixed directions, undefined values, and low reliability \citep{hedge2018reliability}, so structure uses performance scores. The original wager contrast covers only 17 models.

Principal component analysis (PCA) of task correlations, 5,000-permutation parallel analyses, and minimum-residual (MINRES) factor analysis with oblimin rotation describe model profiles and family centroids. Factor intervals refit 2,000 whole-family resamples, and failed, improper, and boundary fits are counted, not replaced.

For prediction we leave out one family at a time. Using only the training families' centroids, we fit standardization, the first principal component (PC1) of the other 12 tasks, and the residualized mean of the same-group tasks, then predict the held-out family. The ridge penalty is fixed at $10^{-6}$, target scores never enter the features, and root-mean-square error (RMSE) weights each family--target cell equally. Gain is $1-\mathrm{RMSE}_{\mathrm{extended}}/\mathrm{RMSE}_{\mathrm{axis}}$, with 20,000 paired family-error-block draws conditional on saved predictions. All 600,600 partitions of the tasks into groups of sizes 3, 3, 3, 2, and 2 give a reference, and PC1+PC2 and two-factor scores are further controls.
\begin{table*}[t]
\centering
\begingroup
\fontsize{8.5}{9.4}\selectfont
\setlength{\tabcolsep}{3pt}
\setlength{\extrarowheight}{1.4pt}
\setlength{\arrayrulewidth}{.25pt}
\begin{tabularx}{\textwidth}{@{}l>{\raggedright\arraybackslash}Xrrrl@{}}
\grouprule
\rowcolor{headgray}
\textbf{Task} & \textbf{Measure} & \textbf{Human} & \textbf{Model} & \textbf{Difference} & \textbf{95\% CI} \\
\grouprule
DS & Forward $-$ backward accuracy & 0.047 & 0.419 & \diffval{0.372} & [0.280, 0.501] \\
\hline
NB & 2-back ordinary non-target false alarms & 0.071 & 0.328 & \diffval{0.257} & [0.159, 0.341] \\
\hline
OS & Letter-position accuracy & 0.763 & 0.582 & \diffval{$-$0.182} & [$-$0.296, $-$0.080] \\
\grouprule
ST & Error increase with conflict & 0.051 & 0.371 & \diffval{0.320} & [0.247, 0.448] \\
\hline
FL & Error increase with conflict & 0.059 & 0.125 & 0.065 & [$-$0.006, 0.137] \\
\hline
GN & No-go false alarms & 0.214 & 0.254 & 0.040 & [$-$0.065, 0.191] \\
\grouprule
CVLT & Recall slope (words per trial) & 1.549 & $-$0.287 & \diffval{$-$1.836} & [$-$2.138, $-$1.476] \\
\hline
DRM & Unwarned new $-$ lure false alarms & $-$0.457 & $-$0.164 & \diffval{0.293} & [0.217, 0.367] \\
\hline
SM & Direct $-$ indirect source accuracy & $-$0.078 & $-$0.027 & \diffval{0.052} & [0.038, 0.062] \\
\grouprule
FB & False-belief accuracy & 0.983 & 0.772 & \diffval{$-$0.210} & [$-$0.297, $-$0.122] \\
\hline
EP & Mental-state accuracy slope & $-$0.017 & $-$0.004 & \diffval{0.013} & [0.008, 0.019] \\
\grouprule
CAL & Type-2 AUROC & 0.714 & 0.573 & \diffval{$-$0.141} & [$-$0.196, $-$0.105] \\
\grouprule
\multicolumn{6}{@{}l@{}}{WAG\quad 37/54 models use constant wagers. No human proportion is available.} \\
\grouprule
\end{tabularx}
\caption{One key endpoint per paradigm in 54 models. Differences are model minus human in original units, and a positive difference does not always mean better. Models receive equal weight, and 95\% intervals use 5,000 whole-family draws with the human mean fixed. Keys were chosen after initial results. Bold differences have intervals that exclude zero.}
\label{tab:h54_behavior_core}
\endgroup
\end{table*}

\section{Behavioral Results}
\label{sec:h54_behavior}
\subsection{Sequence retrieval and updating}
\label{sec:h54_wm}
\begin{figure*}[!t]
\centering
\includegraphics[width=\textwidth,trim=0 16 0 0,clip]{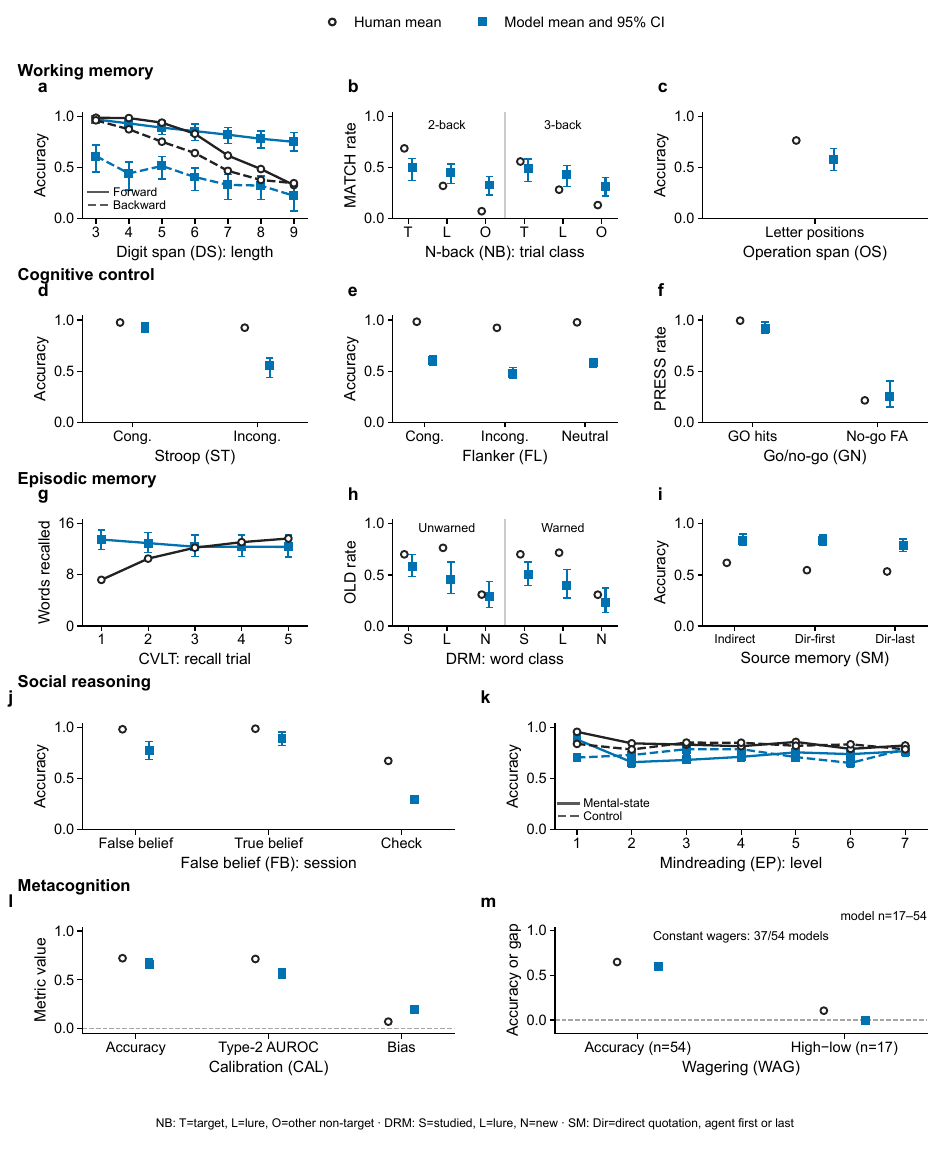}
\caption{Condition-level behavior across all 13 tasks, for 54 models except where a panel states otherwise. Model means weight models equally, and intervals are percentiles of 5,000 whole-family resamples with human means fixed and human uncertainty not shown. Invalid task answers count as wrong. Invalid CVLT steps contribute zero words while valid steps are retained. Human $n$ is DS 188--522, NB 40 paired blocks per load, OS 438, ST and GN 521, FL 519, CVLT 130, DRM 240 unwarned and 197 warned, SM 94--105, FB 49, EP 63, CAL 181, and WAG 40.}
\label{fig:h54-behavior}
\label{fig:h54-memory}
\end{figure*}

Digit-span difficulty runs in the same direction for models and humans, but the sizes differ (Figure~\ref{fig:h54-behavior}). Models do better than humans on long forward lists (length 9, .750 versus .327) and worse on short backward ones (length 3, .611 versus .961), so their forward--backward gap is much larger. With the history visible, these are not capacity estimates. Operation-span accuracy is lower in models overall, but not among length-3 gate passers.

In 2-back, both populations respond MATCH more often to lures than to controls, but models also accept many ordinary non-targets. Their control false-alarm rate exceeds the human rate by .257 [.159,.341], with an excess in 39 of 54 models and in 19 of 23 family means. Deleting any one family leaves .226--.269, and the excess remains without the 16 models that give nearly constant answers (.238, in 33 of 38 models). Models also barely separate targets from lures (hit minus lure rate, .048 versus .367), and only the three native-thinking models exceed humans. At 3-back the control excess is smaller. The shared lure effect therefore does not establish a shared mechanism.

\subsection{Interference and response control}
\label{sec:h54_cc}
Models are far more distracted by the word in Stroop than humans are. Their accuracy falls from .926 on congruent trials to .555 on incongruent ones, against .976 to .925 in humans. On incongruent trials models answer with the word itself .376 of the time, and humans .042. Deleting any one family leaves an excess of .318--.367. It appears in every model family, in 27 of 28 raw family labels, and in the 19- and 51-model panels, but it is smaller in models of 27B or more (.089, in 4 of 9). Flanker interference reliably exceeds the human value only among the 18 models that pass the congruent gate. Go/no-go false alarms do not differ reliably overall, and none of the nine models of 27B or more makes one (humans .214). Text omits the timing and motor demands of these tasks.

\begin{table*}[t]
\centering
\begingroup
\fontsize{8.5}{9.4}\selectfont
\setlength{\tabcolsep}{3pt}
\setlength{\extrarowheight}{1.4pt}
\setlength{\arrayrulewidth}{.25pt}
\begin{tabularx}{\textwidth}{@{}>{\raggedright\arraybackslash}p{10mm}>{\centering\arraybackslash}p{15mm}>{\raggedleft\arraybackslash}X>{\raggedright\arraybackslash}p{36mm}>{\raggedleft\arraybackslash}X>{\raggedleft\arraybackslash}p{28mm}@{}}
\grouprule
\rowcolor{headgray}
& \multicolumn{3}{c}{\textbf{Gate passers}} & \multicolumn{2}{c@{}}{\textbf{Models $\geq$27B ($n=9$)}} \\
\cmidrule(lr){2-4}\cmidrule(l){5-6}
\rowcolor{headgray}
\textbf{Task} & \textbf{$n$} & \textbf{Difference} & \textbf{95\% CI} & \textbf{Difference} & \textbf{Same sign} \\
\grouprule
DS & 53 & \diffval{0.378} & [0.289, 0.503] & 0.151 & 5/9 \\
\hline
NB & 32 & \diffval{0.266} & [0.155, 0.336] & 0.351 & 7/9 \\
\hline
OS & 33 & 0.066 & [$-$0.007, 0.115] & 0.143 & 2/9 \\
\grouprule
ST & 51 & \diffval{0.348} & [0.264, 0.488] & 0.043 & 3/9 \\
\hline
FL & 18 & \diffval{0.319} & [0.176, 0.467] & 0.100 & 4/9 \\
\hline
GN & 45 & $-$0.049 & [$-$0.131, 0.078] & $-$0.214 & 0/9 \\
\grouprule
CVLT & 46 & \diffval{$-$1.954} & [$-$2.306, $-$1.618] & $-$1.549 & 9/9 \\
\hline
DRM & 21 & 0.058 & [$-$0.113, 0.175] & 0.110 & 6/9 \\
\hline
SM & 52 & \diffval{0.050} & [0.036, 0.061] & 0.098 & 9/9 \\
\grouprule
FB & 49 & \diffval{$-$0.149} & [$-$0.214, $-$0.092] & $-$0.078 & 2/9 \\
\hline
EP & 28 & \diffval{0.015} & [0.011, 0.022] & 0.013 & 8/9 \\
\grouprule
CAL & 30 & \diffval{$-$0.108} & [$-$0.161, $-$0.052] & $-$0.026 & 6/9 \\
\grouprule
\multicolumn{6}{@{}l@{}}{WAG\quad Constant wagers in 15/24 gate passers and 4/9 models $\geq$27B.} \\
\grouprule
\end{tabularx}
\caption{Gate and size comparisons with the endpoints and human references of Table~\ref{tab:h54_behavior_core} and the gates of Section~\ref{sec:h54_gates}. Differences are model minus human, and intervals follow Table~\ref{tab:h54_behavior_core}. The nine models of 27B or more span six families without native thinking, and Section~\ref{sec:h54_gates} gives their NB, ST, and CAL intervals. Same sign counts the large models whose gap has the sign in Table~\ref{tab:h54_behavior_core}. Bold marks gate-passer differences whose intervals exclude zero.}
\label{tab:h54_behavior_gates}
\endgroup
\end{table*}

\subsection{List retrieval and source attribution}
\label{sec:h54_em}
In CVLT, models recall most of the list from context on the first trial (13.50 of 16 words versus 7.20) and do not improve over trials (slope $-.287$ versus 1.549). They also make more recognition false alarms (.281 versus .043) and cued intrusions (19.17 versus 1.08). The materials stay visible and delays are only stage labels, so the model curve is not a learning or retention estimate (Figure~\ref{fig:h54-memory}).

In DRM, models accept fewer related lures than humans without a warning (.451 versus .765), and a warning lowers lure false alarms by about the same amount in models ($-.053$) and humans ($-.048$), although the human warning groups are different people. Among the 21 DRM gate passers, the difference in the new-minus-lure contrast shrinks from .293 to .058 [$-.113,.175$]. Source-memory accuracy is higher in models in all three quotation conditions, and direct speech is slightly less accurate than indirect speech in both populations.

\subsection{Belief and recursive mindreading}
\label{sec:h54_tom}
Humans are near ceiling on false-belief (.983) and true-belief stories (.988). Models reach .772 on false-belief and .892 on true-belief stories and do much worse on check questions (.296 versus .673). Their main-item accuracy also shifts with presentation (.778--.877), and 16 to 34 models answer all 14 main items correctly, depending on presentation. In recursive mindreading, model accuracy declines less steeply across orders than human accuracy. Mental-minus-control contrasts change direction across orders, so the flatter slope does not show better recursive reasoning.

\subsection{Confidence and wagering}
\label{sec:h54_mc}
Models are less accurate than humans on the calibration items (.659 versus .722) yet more confident (.859 versus .793), giving a larger source-defined bias (.200 versus .071). Their confidence also separates correct from incorrect answers less well. Type-2 AUROC is lower (.573 versus .714) in 47 of 54 models, in every model family, in both answer orders, and in the 19- and 51-model panels, and deleting any one family leaves a gap of $-.164$ to $-.132$. Higher confidence and lower accuracy coexist in only 21 models, so the averages do not describe each model.

Token probabilities carry more of this signal than verbal reports. We reran the 51 direct-answer models with the same Ollama build and weights and read the probabilities of the two options at the first answer position, renormalized over the two among the top 20 tokens \citep{kadavath2022language,steyvers2025large}. Rerun responses matched the formal run on more than 95\% of steps. In the 41 models with both options among the top 20 on at least 40 of 80 items (a post hoc threshold), token AUROC is .763, above matched-item verbal AUROC (.569) in 39 of them and slightly above humans, by .049 [.021,.067]. Across 48 computable models it is .705, close to humans. The two measures correlate weakly within models (median $\rho=.16$).

Wagering accuracy is slightly lower in models (.595 versus .650), and most models do not use the wager at all. Thirty-seven always choose the same wager (29 low, eight high), across 19 of 23 families. This includes accurate classifiers. Gemma3 27B is 86.7\% accurate but always wagers low. High-minus-low accuracy is undefined for these models, not zero. The formal run asked for a wager of 1 or 2 sweets without counterbalancing the labels. Under both A/B wager-label mappings in the rerun, 30 of 51 models choose the same letter on at least 90\% of wagers, and only seven switch letters with the amount on at least 80\%. These 30 include 21 of the 36 rerun models with constant formal wagers, so constant wagering mostly tracks response labels.

\begin{figure*}[!t]
\centering
\setlength{\abovecaptionskip}{3pt}
\includegraphics[width=\textwidth,trim=0 13 0 0,clip]{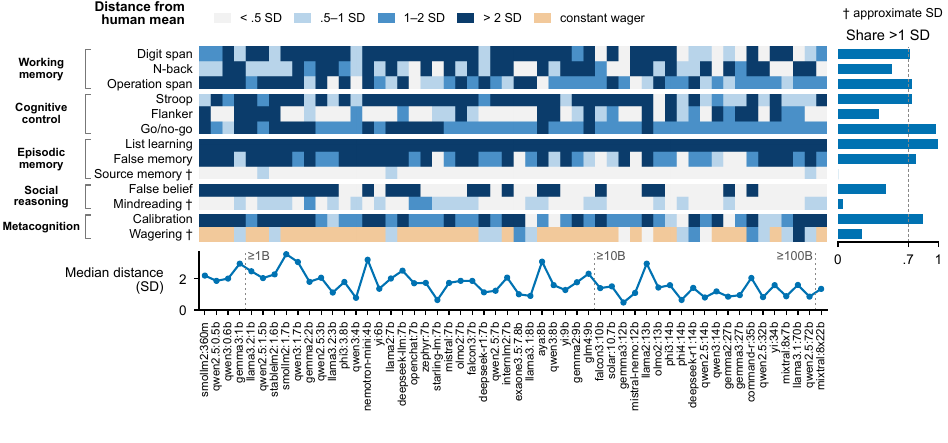}
\caption{Each model's distance in human SDs from the human mean on each key endpoint (Table~\ref{tab:h54_behavior_core}, high-minus-low accuracy for WAG), models ordered by size. Bars give the share over 1 SD (WAG among 17 models with a value), and the bottom panel each model's median. $\dagger$ SDs are approximate (Section~\ref{sec:h54_analysis}).}
\label{fig:h54-closeness}
\end{figure*}

\subsection{Minimal-performance gates and scale}
\label{sec:h54_gates}
Gates ask whether a difference survives among models that meet a minimal level on the same task. They require above-chance accuracy (one-sided binomial $p<.05$) on congruent FL and ST, EP control order 1, and overall SM, CAL, and WAG, hits above false alarms (one-sided Fisher $p<.05$) on NB 2-back, GN, and unwarned DRM, or at least half correct on DS and OS length 3, CVLT trial 1, and FB true belief. Congruent ST and FL can be passed by reading words or flankers, so gates do not establish comprehension. Eleven tasks share gate and endpoint trials, and gates built from separate trials give similar gaps (Flanker .341--.352, OS .057--.070, CAL $-.140$ to $-.061$).

Among gate passers, the n-back false-alarm excess and the lower AUROC remain (Table~\ref{tab:h54_behavior_gates}). Passing a gate does not mean performing well, though. Of the 32 n-back gate passers, 19 have balanced accuracy below .60, and only three, all native-thinking, reach .80. Constant wagers persist in 15 of 24 above-chance grammar classifiers, including four of the five scoring at least .80.

Larger models narrow some gaps but not all. Among the nine models of 27B or more, the n-back excess remains (.351 [.095,.644], 7 of 9). Their Stroop interference (.043 [$-.031,.142$]) and AUROC ($-.026$ [$-.140,.101$]) gaps are smaller, with wide intervals. Family-clustered mixed models with log parameters and native thinking give an AUROC slope of .123 per tenfold increase [.060,.187]. The false-alarm slope (.167) is not reliably nonzero (likelihood-ratio $p=.06$, Wald lower bound .0005), nor is the interference slope.

On seven of the 13 tasks, more than 70\% of models lie over one human SD from the human mean, and this distance does not shrink consistently with size (Figure~\ref{fig:h54-closeness}).

\section{Score Structure}
\label{sec:h54_fullbattery}
\label{sec:h54_structure}
Secondary structure analyses use 54 models in 23 families with no missing cells. Joint human cohorts with grouping contrasts are SRO (five tasks, 345 people) and Ball (three measures, 409).

\paragraph{Shared variation and scale.}
Principal-component and common-factor parallel analyses both retain two dimensions, in models and in family centroids. Model-level PC1 explains 39.23\% of standardized variance and correlates with log parameter count ($r=.797$, and .839 within multi-model families), as in benchmark factor and scaling studies \citep{burnell2023revealing,ilic2024evidence,ruan2024observational}. Residualizing every task on log size lowers PC1 to 27.00\% but still leaves two dimensions, as does dropping WAG, whose constant wagers score .5 (post hoc).

The first rotated factor loads broadly across origin groups. The second rotated factor contrasts NB and FL with WAG and CVLT. It is unstable under family resampling (median congruence .938 [.421,.995]). It has no reliable association with three prespecified response-bias indicators, NB control false alarms, constant wagers, and high-wager rate (Spearman $\rho$ from $-.23$ to $.12$, all intervals spanning zero). Two secondary associations, with GN false alarms ($-.47$) and CVLT false alarms ($-.42$), are negative, and none establishes a general response-bias reading. The second component's eigenvalue only modestly exceeds the permutation reference (2.22 versus 1.82). Adding PC2 to PC1 gives no reliable prediction gain (3.17\% [$-1.53,15.73$]\%). With the 12 key endpoints instead, parallel analysis retains one dimension and the five groups give no reliable gain ($-2.02$\% [$-5.10,.48$]\%).

\paragraph{Predefined groups and random groups.}
Adding same-group information to a PC1 baseline in family-held-out prediction lowers RMSE from 1.563 to 1.515 (3.02\% [$-5.85,6.01$]\%). Of all 600,600 same-size random groupings, 14.62\% do at least as well. Within- and between-group correlations are similar ($\delta=-.021$ [$-.085,.066$]). Model split or form reliability is .869--.999 except for WAG (.515). Reliability correction leaves $\delta^*=-.011$ [$-.078,.075$]. The predefined groups therefore show no reliable predictive advantage, which is not evidence that they are equivalent to random groups.

A post hoc positive control asks whether this test can detect grouping of human size, simulating 100 score matrices per group strength. With population $\delta$ near the SRO estimate of .16, at most 5\% of random groupings match the predefined one in 97 of the 100 matrices, although the gain interval excludes zero in only 26. Two alternative groupings of the same sizes, chosen after the formal results, give no reliable gain either. Grouping by response format (span, detection, labelled choice, mind stories, confidence) gives 4.63\% [$-6.58,8.52$]\%, and a grouping chosen within each fold from training-family correlations gives 0.80\% [$-4.10,11.68$]\%. The test concerns these 13 text-adapted paradigms, so it cannot show that models lack such abilities.

\paragraph{Human cohorts.}
In SRO, the human grouping contrast exceeds the models' ($\delta_H=.158$, $\delta_M=-.022$), but the difference, .180 [$-.102,.371$], spans zero, also after reliability correction. Model reliabilities are at least .972, so the weak model working-memory correlation (.076 versus .317 in humans) is not explained by low reliability. In Ball, the comparison depends on DRM scoring. Native false-alarm scoring gives a difference of .249 [.064,.402], but both groups' contrasts are negative. With pooled 56-probe accuracy the difference spans zero. Pooled balanced accuracy gives .493 [.195,.704], with positive intervals in both warning groups (.423 [.133,.651] without and .561 [.231,.810] with a warning). In two further cohorts covering CVLT and FB, model correlations exceed human ones in five of six task pairs, reliably only for DS--ST (.44 in models versus .06 in humans). Bradford measures are adjusted for age.

\section{Discussion}
\paragraph{Effect directions and errors.}
Sharing a human effect direction does not mean making human errors. In n-back the lure effect sits on top of many false alarms to ordinary letters, and in Stroop models often answer with the word instead of the color label. These task-specific patterns recur across many families, but they need not share one cause.

\paragraph{Confidence reports.}
Most models' confidence reports separate correct from incorrect answers less well than humans' do. Their token probabilities discriminate about as well as human reports, so the gap lies in the verbal report rather than in a missing correctness signal \citep{kadavath2022language,tian2023just,xiong2024can,steyvers2025large}. Neither measure shows that models use confidence to control behavior \citep{kumaran2026causal}. Constant wagers mostly follow response labels, so they reflect label preference more than confidence.

\paragraph{Gates and scale.}
Several differences persist above minimal-performance gates. Larger models close some gaps, but distance from humans does not shrink consistently with size (Figure~\ref{fig:h54-closeness}). The gaps describe this panel of models, not fixed properties of language models.

\paragraph{Structure and cognitive interpretation.}
Model scores share a broad, scale-related dimension and a weaker, unstable second one not reliably tied to response bias. Grouping tasks by origin adds no reliable predictive value, so the labels describe paradigm origins, not established model abilities.

\paragraph{Implications for evaluation.}
Human comparisons should report errors by condition, not only effect directions. Confidence should be read from verbal reports and token probabilities, which can disagree. Wagers need counterbalanced labels, or they may measure label preference. Per-model distances from humans show what panel means hide. 

\section{Conclusion}
Across 54 models and 13 paradigms, models often share human effect directions while differing in errors and confidence reports. These differences span many families, and larger models are not consistently closer to humans. Predefined cognitive groups lack reliable predictive advantage over random groupings of the same size, and alternative groupings add no reliable gain either, so behavioral measurements should be kept separate from cognitive interpretations. CogBattery lets new models be tested against the same condition-level human references. Model replies will be released upon publication, apart from those limited by material licenses.

\section*{Limitations}
No humans were tested on the text versions, so human--model differences combine adaptation and population differences. Each model gives one temperature-zero response per step, so response distributions are not estimated. Closed models were not tested. Joint human cohorts cover ten tasks, so CAL, EP, and WAG have no human structural comparison, and the CNP and Bradford tasks match the model tasks only at task-family level. Published materials may occur in training data, and we ran no contamination test. Standardized distances depend on human variability, so tasks with large person-level spread (SM, EP) or ceiling scores in both groups (FB) look closer.

Of the 104 intervals across all endpoints that exclude zero with human means fixed, 95 still do when human sampling error is added. The SM difference in Table~\ref{tab:h54_behavior_core} is one that no longer does. The small five-group gain depends on WAG, CAL, and the deepseek-r family. Whether these patterns hold for other paradigms needs a prospective test.

\FloatBarrier
\clearpage
\bibliography{references,human_references,hu2025nback,h54_references}
\end{document}